# Ontology in Holonic Cooperative Manufacturing: A Solution to Share and Exchange the Knowledge

Ahmed R.Sadik[1,2], Bodo Urban[1,2]

[1]University of Rostock, Universitätsplatz 1, 18055, Rostock, Germany
[2]Fraunhofer Institute for Computer Graphic Research IGD,
Joachim-Jungius-Str. 11, 18059, Rostock, Germany
{ahmed.sadik, bodo.urban}@igd-r.fraunhofer.de

**Abstract.** Cooperative manufacturing is a new trend in industry, which depends on the existence of a collaborative robot. A collaborative robot is usually a light-weight robot which is capable of operating safely with a human co-worker in a shared work environment. During this cooperation, a vast amount of information is exchanged between the collaborative robot and the worker. This information constructs the cooperative manufacturing knowledge, which describes the production components and environment. In this research, we propose a holonic control solution, which uses the ontology concept to represent the cooperative manufacturing knowledge. The holonic control solution is implemented as an autonomous multi-agent system that exchanges the manufacturing knowledge based on an ontology model. Ultimately, the research illustrates and implements the proposed solution over a cooperative assembly scenario, which involves two workers and one collaborative robot, whom cooperate together to assemble a customized product.

## 1    Introduction: Challenges in Cooperative Manufacturing

Cooperative and collaborative manufacturing are new terms in industry, which are usually mixed with each other. The reason behind this confusion is that both terms involve a cooperative/collaborative robot (cobot). However, a slight difference can distinguish between the two terms. In cooperative robotics, both the worker and the cobot are sequentially performing separate tasks over the same product in the same-shared workspace. But in collaborative robotics, they perform a shared task simultaneously [1]. The goal of the cooperative manufacturing is to combine the advantages of both the cobot and the worker at the same time, to afford a new intelligent manufacturing technique. However, with this new technique appears new challenges [2].

Fig. 1 shows the challenges in cooperative robotics. These challenges are beyond the physical safety of the worker during the cooperation, which already gained a great focus in robotics research [3]. The first challenge after ensuring the worker physical safety, is to achieve the physical interaction between the cobot and the worker. The physical interaction can be achieved via a group of sensory and User Interface (UI) devices. The sensors and the UI convert the worker actions into input information.

Exchanging this information with the cobot and the other manufacturing components is the second challenge in collaborative manufacturing. The third challenge is to build the manufacturing knowledge from this exchanged information. Thus, this knowledge is used to plan and control the manufacturing activities. Ultimately, comes the final challenge of collecting the knowledge from all the manufacturing components and reason it to provide collective decisions.

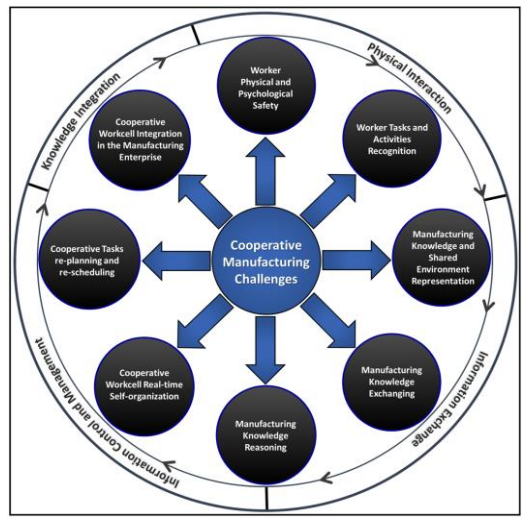

**Fig. 1.** Cooperative Robotics Challenges

## 2  Problem Formulation

This article is focusing on two important challenges in cooperative manufacturing, which are discussed as below:

Manufacturing knowledge and environment representation: the abstract representation of the facts and the objects within the cooperative workcell is an essential necessity to establish a successful cooperation. Thus, in order to construct the cooperative manufacturing concept, a proper approach is required to describe the shared environment between the worker and the cobot including themselves. Also, the other production components such as the product parts and the manufacturing tools. Descriptive meta-data representation is essential to give a meaning of the objects, tasks, and operations within the cooperative workcell. Structural meta-data are required to structure bonds between the objects to compound new descriptive meta-objects, this also can be done by associating new attributes to an existing object. Also, the structure meta-data can be used to define the parent-child relation among the objects. Finally, administrative meta-data is required to control the cooperative task assignment and the cooperation planning, management, and execution.

Manufacturing knowledge exchanging: the knowledge is useless, unless it is exchanged and being used in reasoning. The knowledge cannot be shared unless a proper mean of communication allows its exchange. Due to the fundamentals of the Human-Robot Interaction (HRI), the information exchange is as much important as the physical safety. The manufacturing information and communication control system can be seen as the nervous system which connects the cobot and the worker together in one body. In the absence of the proper manufacturing control solution, the maximum usability of a cobot can never be reached, because it loses its real value as a smart tool. The challenge from the research point of view is to provide a communication language which can express the relations among the cooperative workcell components. Moreover, the actions and operations that can be performed by these components. Putting into consideration that this communication language should be human readable and in the same time can be processed by the machine (i.e., the cobot). The challenge from the technical point of view is to provide a communication mean that grantees the industrial connectivity (i.e., interoperability).

## 3  Solution Preliminaries

### 3.1     Autonomous Agent Communication

A software agent is a computer system situated in a specific environment that is capable of performing autonomous actions in this environment in order to meet its design objective. An agent is autonomous by nature; this means that an agent operates without a direct intervention of the humans, and has a high degree of controlling its actions and internal states [4]. In order to achieve this autonomy, an agent must be able to fulfil the following characteristics:
- Social: can interact with other artificial agents or humans within its environment in order to solve a problem.
- Responsive: capable of perceiving its environment and respond in a timely fashion to the changes occurring in it.
- Pro-active: able to exhibit opportunistic, goal directed behavior and take initiative.

Conceptually, an agent is a computing machine which is given a specific problem to solve. Therefore, it chooses certain set of actions and formulates the proper plans to accomplish the assigned task. The set of actions which are available to be performed by the agent are called a behaviour. The agent behaviours are mainly created by the agent programmer. An agent can execute one or more behaviour to reach its target. The selection of an execution behaviour among others would be based on a certain criteria which has been defined by the agent programmer. Building an execution plan is highly depending on the information which the agent infers from its environment including the other agents. A Multi-Agent System (MAS) is a collective system composed of a group of artificial agents, teaming together in a flexible distributed topology, to solve a problem beyond the capabilities of a single agent [5].

Java Agent DEvelopment (JADE) is a distributed middleware framework that can be used to develop an MAS as it is shown in Fig. 2 [6]. Each JADE instance is an independent thread which contains a set of containers. A container is a group of agents run under the same JADE runtime instance. Every platform must contain a main container. A main container contains two necessary agents which are: an Agent

Management System (AMS) and a Directory Facilitator (DF). AMS provides a unique Identifier (AID) for every agent under its platform to be used as an agent communication address. While the DF announces the services which agents can offer under its platform, to facilitate the agent services exchange, so that every agent can obtain its specific goal. JADE applies the reactive agent architecture which complies with the Foundation for Intelligent Physical Agent (FIPA) specifications [7]. FIPA is an IEEE Computer Society standards organization that promotes agent-based technology and the interoperability of its standards with other technologies. JADE agent use FIPA-Agent Communication Language (FIPA-ACL) to exchange messages either inside or outside its platform

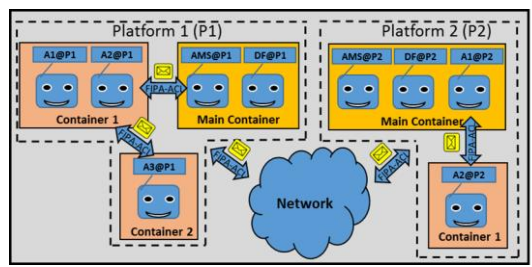

**Fig. 2.** JADE Framework - an example

JADE agent follows FIPA communication model stack which is very similar to the TCP/IP model as it can be seen in Fig. 3. Generally, a communication architecture model is a stack of layers, each layer specifies a set of data flow, information management, and communication control standard protocols. The purpose of a communication architecture model is to exchange the services and the information among the computing machines within a wired or a wireless network. FIPA communication model has in total 9 layers. The first 3 lower layers are exactly the same as in the TCP/IP model, while the upper 6 layers equal together to the application layer in the TCP/IP model. A brief description of FIPA communication model can be seen as below:

1- Physical Interaction Layer: this layer is linked directly to the actual network medium which is a part of the computing machine hardware. Standard IEEE 802.3-Ethernet and IEEE 802.11-Wireless Ethernet protocols are responsible for sending/receiving the network data packets.
2- Communication Interaction Layer: this layer uses the Internet Protocol (IP) to pack/unpacking the data into/from the IP datagrams. Also it provides the right network rout and IP-address for the datagrams.
3- Transport Layer: the transport layer makes sure that the IP datagrams are delivered from the source to the destination and vice-versa. Two famous communication control protocol can be used in this layer, which are Transmission Control Protocol (TCP) and User Datagram Protocol (UDP). TCP deploys a fixed length datagram via peer to peer connection with a reliable acknowledgment of packets delivery. While UDP deploys a variable length datagrams via peer to peer or broadcasting connection with no need of packets delivery acknowledgment.

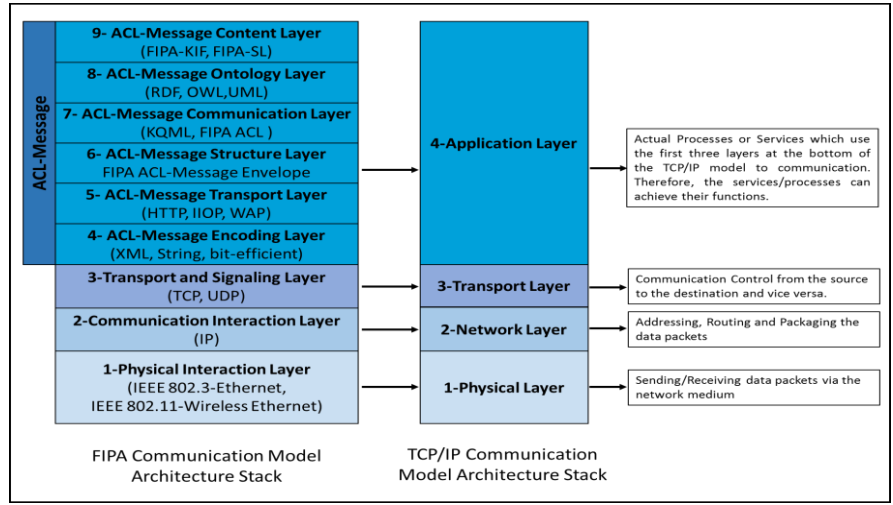

**Fig. 3.** An analogy between FIPA and TCP/IP Communication Model Architectures

4- FIPA ACL-Message Encoding Layer: this layer specify the message presentation standard format. Extensible Markup Language (XML) can be used to XML is the most famous standard to represent a structured electronic message. Other FIPA standard format such as String and bit-efficient can be also used to represent and ACL- Message.

5- FIPA ACL-Message Transport Layer: Message Transport Layer (MTP) is implementing standard application protocols which deal with the web applications such Hypertext Transfer Protocol (HTTP), Internet Inter-ORB Protocol (IIOP), and Wireless Application Protocol (WAP). These protocols are designed to facilitate the distributed collaborative information exchange via request-response interaction in client/server computer model.

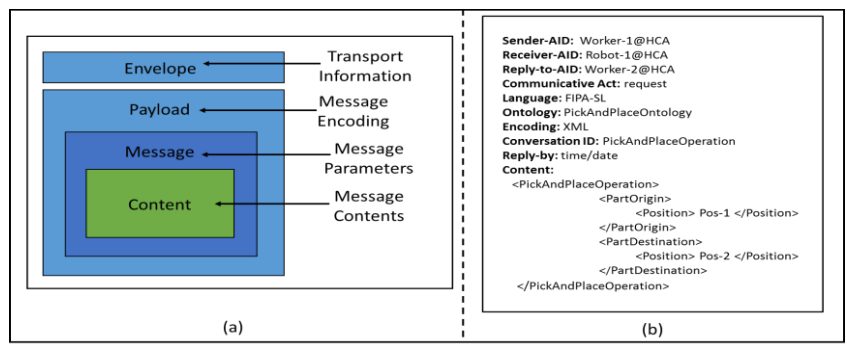

**Fig. 4.** (a) ACL-Message Structure [Fip15] – (b) ACL-Message Example

6- FIPA ACL-Message Structure Layer: an ACL-Message has a standard structure which is necessary to encode the message. An ACL-Message structure depends on the upper 6 layers of FIPA communication model. A schematic of an ACL-Message structure can be seen in Fig. 4-a and an example of the message can be seen in Fig. 4-b. The fields of an ACL are coded via the agent programmer and they can vary due to the application and implementation approach.
7- FIPA ACL-Message Communication Layer: this layer is very important as it defines a set of terms which are mandatory to build an ACL-Message. FIPA-ACL is the official FIPA standard communication language which extends the features of Knowledge Query and Manipulation Language (KQML). For example, the communication act is a mandatory filed which has a standard values which are defined by the FIPA. The value of a commutation act can be seen as a specific type of the message which must express the purpose of the message as it is shown in Table 1. A conversation-ID field is mandatory field as well, however its value is up to the agent coder. A conversation-ID is very important concept to track a thread of messages in a specific conversation during the interaction.

Table 1. FIPA ACL Communication Acts

| Communication Act | Purpose |
| --- | --- |
| propose, accept-proposal, reject-proposal, cfp | negotiation |
| request, request-when, query-if, query-ref | requesting information |
| confirm, disconfirm, inform, inform-if, inform-ref | passing information |
| agree, refuse, cancel, subscribe | performing actions |
| not-understood, failure | error handling |
| propagate, proxy | message referencing |

8- FIPA ACL-Message Ontology Layer: an ACL-Message is flexible to afford different levels of conversation complexity. A String based massage is the simplest form of conversation among FIPA agents. However in a complex agent conversation scenario, an ontology-based content will be the proper conversation method. In ontology-based ACL-Message, some specific ontology languages are used to structure the message. FIPA support Resource Description Framework (RDF), Web Ontology Language (OWL), and Unified Modeling Language (UML) which are the most common languages to code an ontology.
9- FIPA ACL-Message Content Layer: FIPA goes beyond a conceptual modelling via an ontology. An ontology model is often used by an agent to represent its own environment. However, agents are meant to exchange and process information in order to achieve MAS cooperation concept. Accordingly, reasoning an ontology based message is an essential task of an autonomous agent. Therefore, FIPA supports logic based ontology model via two languages which are FIPA Knowledge Interchange Format (FIPA-KIF) and FIPA Semantic Language (FIPA-SL). FIPA-KIF and FIPA-SL languages allow to model the system knowledge based on first order logic and commonly used in AI. Therefore, using a logic based ontology model enables a reasoning software to process and analysis the knowledge of this model.

## 3.2 Ontology Concept in Cooperative Manufacturing

In philosophy, an ontology aims to study the nature of the existence and reality to form tight relations among the beings whom are part of this reality. The word ontology comes from combining two Greek words "Onto" which means a Being and "logia" which means a Divine Origin. Thus, ontology can be translated literally into the origin of the being. However, the concept of ontology has been frequently used in computer science in association with the AI to define the relation among the software objects. Although there is no a specific definition of the word ontology within the computer science filed, the following definitions shall draw a complete understadning of the meaning of ontology [8]:

- An ontology defines the basic terms and relations comprising the vocabulary of a topic area as well the rules for combining terms and relations to define extensions for this vocabulary [9].
- An ontology is a formal, explicit specification of a shared conceptualization [10].
- An ontology is a logical theory accounting for the intended meaning of a formal vocabulary [11]. i.e., "it is a commitment to a particular conceptualization of the world".
- An ontology provides the meta-information to describe the data semantics, represent knowledge, and communicate with various types of entities (e.g., software agents and humans) [12].
- An ontology can be described as the means of enabling communication and knowledge sharing by capturing a shared understanding of terms that can be used by both the humans and the machine software [13].

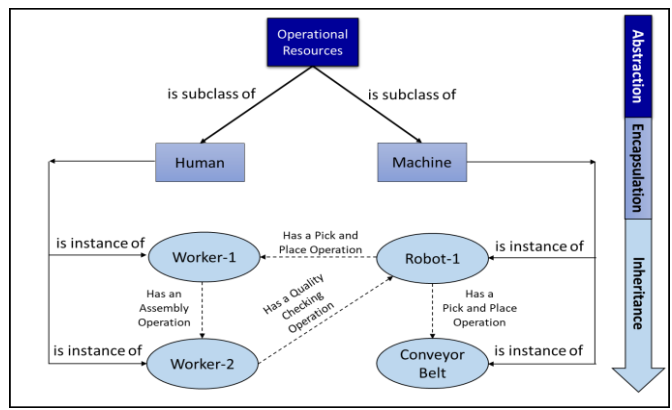

**Fig. 5.** An Ontology Example (Cooperative Operational Resources)

The previous definitions of ontology can derive a precise definition of the ontology within the scope of cooperative manufacturing. Thus, an ontology is a conceptual tool to represent and create a common understanding for the manufacturing workcell entities. Furthermore, this common understanding would enable to exchange, reuse and extend the manufacturing knowledge. An ontology based model is extending the

Object Oriented Programming (OOP) concepts by defining relations among the objects. Fig. 5 illustrates an ontology example of a cooperative workcell which achieves the OOP concepts that can be summarized as the following:

- Abstraction: is a generalization process which can differ due to the context and the purpose of the code. In another words, abstraction is the process of exposing an object parameters which are general and can be used by all the other objects within the software domain.
- Encapsulation: can be seen as the opposite of abstraction. As it is simply means to hide an object parameters which can only be used by the object, to fit the context of the software domain.
- Inheritance: is the idea of having different instances of the same object with different parameters value.

### 3.3  Agent Ontology-Based Interaction

An ACL-Message is the main tool which an agent uses to interact among the other agents to achieve the cooperation goals. The power of an autonomous agent which distinguishes it from a software object locates in its ability to interact rather than applying static algorithms. An agent interaction is a form internal conversation where a group of agents negotiate to reach a mutual acceptable agreement on a certain matter [14]. An agent interaction can be based on cooperative or competitive negotiation. Competitive negotiation takes place when the agents have conflicted local goals, which every agent races to achieve with a minimum cost function. Cooperative negotiation takes place when the agents have a common global goals, which they try to achieve with a minimum cost function. Competitive and cooperative negotiation can exist along together in the same MAS [15].

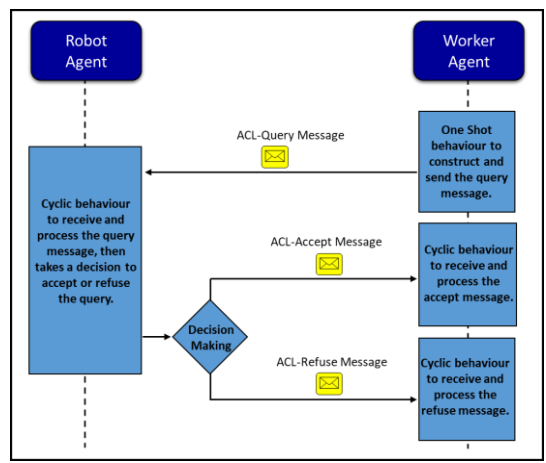

**Fig. 6.** Autonomous Agents Interaction Pipe Line Example

From a philosophical point of view, an agent is similar to the human being who has a group of behaviours and uses one or a group of them concurrently to respond to the environment situations. From the software point of view, a behaviour is an event handler routine which the agent uses to modify its parameters and negotiate with other agents. JADE offers different behaviours which are used to build an agent, the most common behaviours are discussed as below:

- One Shot Behaviour: is executed once when it is called by the agent, then it ends. It is very useful to trigger an event and to send an ACL-Message.
- Cyclic Behaviour: stays active as long as the agent is alive. It is very useful to receive a message with specific conversation-ID or communication act.

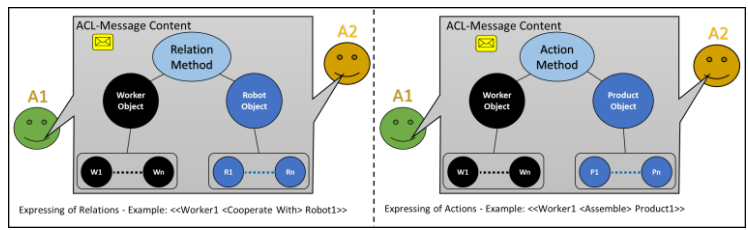

**Fig. 7.** Agents Communication via Ontology

A software agent goes many steps farther than the software object by providing new concepts and tools such as the behaviours and the ACL-Message. One of the main roles of a behaviour is to control the follow of the ACL-Messages. The summation of an agent behaviour and an ACL-Message should infer the meaning of an agent interaction. The research in [16] states that "agent interaction is more powerful concept than ordinary algorithm". The main reason behind the previous statement that an interaction is providing the capability of learning from the input/output history tracking and changing the environment model based on this knowledge. While, algorithm is static and meant to produce an output due to a specific input [17]. Furthermore, using the concept of interaction via the agent behaviours and ACL-Messages, gives the agent the sense of intelligence by taking its own decisions based on the interaction context as it can be seen in Fig. 6.

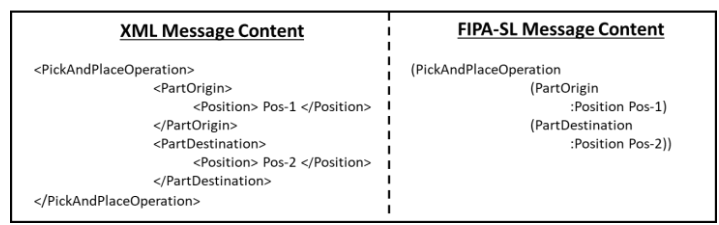

**Fig. 8.** XML Syntax versus FIPA-SL Syntax

The representation of the content field of an ACL-Message is String data type in case of simple agent conversation. However in a complex agent conversation, it is not only a must to representing the software objects, but also the relations among these object and the actions that they can perform as it is shown in the example in Fig. 7.

Therefore, an ontology-based content will be the proper conversation method. XML or FIPA-SL languages are commonly used to represent an ACL-Message content. XML and FIPA-SL are slightly different in their syntax as it can be seen in Fig. 8. However, representing the ACL-Message in FIPA-SL is preferable in case of further reasoning of the message is needed based on logic theory. The representation of an ACL-Message can be also as a sequence of Bytes which is a light-weighted method from the programming point of view, but not perforable at all as it is non-human readable language, thus it cannot be debugged [18].

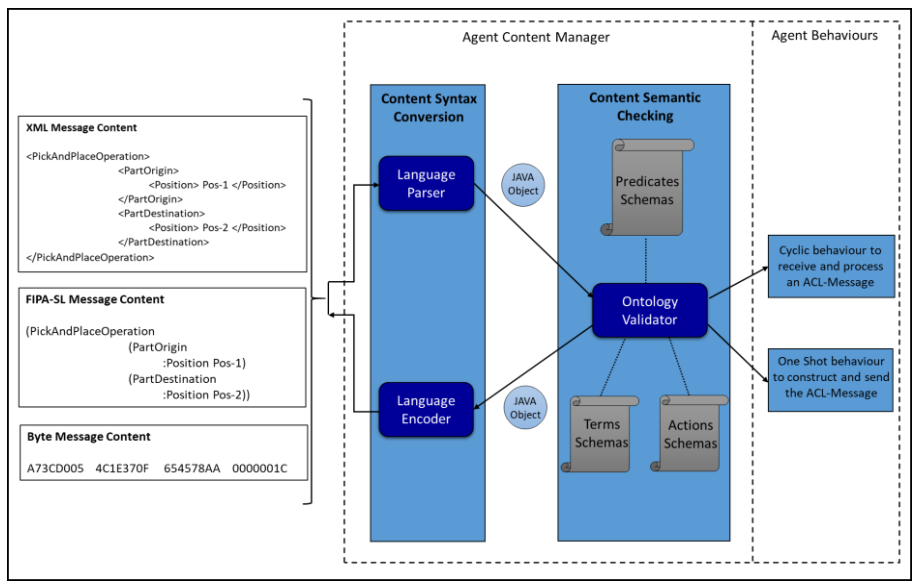

**Fig. 9.** Sending/Receiving Mechanism of an ontology-based ACL-Message

One of the advantages of using the ontology is the mutual understanding among the software agent. In order to bring this mutual understanding among the agents, they should communicate via the same syntax and semantics. JADE agents are written as a JAVA code which obligate them to process JAVA objects. Therefore, during sending an ontology-based message, the sender must convert its internal representation of the message from JAVA syntax to the appropriate syntax of the corresponding content language (i.e. XML, FIPA-SL, Bytes). However, before the syntax conversion, semantic checking should be fulfilled. Thus, an agent must check and validate an ontology-based ACL-Message against the domain semantics before converting it to the proper syntax. Ontology schemas are the approach which an agent uses to express the domain semantics. JADE agents define three types of schemas which can be summarized as the following:
- Terms: a schema which contains expressions that indicate entities that exist in the agent domain and that agents may reason about. Terms can be seen as primitives which are atomic data types such as strings or integers, and concepts which are complex structure such as objects.

- Predicate: a schema which contains expressions that describe the status of the agent domain and the relationships between the concepts.
- Action: a schema which contains expressions that describe routines or operations which can be executed by an agent.

During an ontology-based ACL-Message reception, the opposite mechanism of the sending operation must be fulfilled. Therefore, the same schemas of the sender must be accessible by the receiver in order to interpret the message content. An illustration of an ontology-based ACL-Message interaction is shown in detail in Fig. 9.

## 4  Solution Concept: Holonic Control Architecture

In the late sixties, the term holon has been introduced for the first time by philosopher Koestler [19]. Koestler developed the term as a basic unit in his explanation of the evolution of the biological and social structures. Based on his observations that organisms (e.g., biological cells) are autonomous self-reliance units, which have a certain degree of independent control of their actions, yet they still subject to higher level of control instructions. His conclusion was that any organism is a whole "holos" and a part "on" in the same time, which derived the term holon [20]. The concept of holon has been adopted in the early nineties by the intelligent manufacturing systems (IMS) consortium, to define a new paradigm for the factory of the future. The following terminologies has been defined by the IMS to provide a better understanding of the Holonic Control Architecture (HCA):

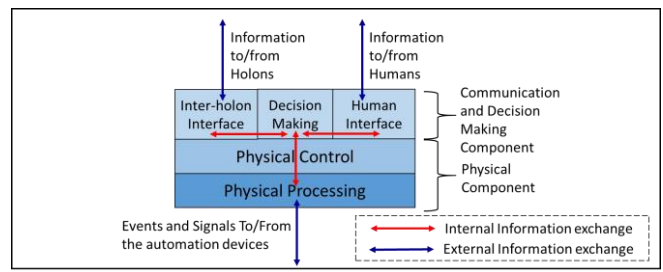

**Fig. 10.** Holon Conceptual Model

- Holon: an autonomous cooperative building block of the manufacturing system that can be used to transform, transport, store and/or validate the information and the physical signals [21]. Fig. 10 shows in detail the holon conceptual model.
- Autonomy: the capability of the holon to create and control the execution of its own plans and/or strategies.
- Cooperation: a process whereby a set of holons develop mutually acceptable plans and execute these plans together.
- Holarchy: a system of holons which cooperate to achieve a goal or objective. The holarchy defines the basic rules for cooperation of the holons and thereby limits their autonomy.

The HCA is basically a distributed control and communication topology which divides the manufacturing process tasks and responsibilities over three basic holon categories [22] which are as following:

- Product Holon (PH): is responsible for processing and storing the different production plans required to insure the correct manufacturing of a certain product.
- Order Holon (OH): is responsible for composing, managing the production orders. Furthermore, in a small-scale enterprise, it should assign the tasks to the present operating resources and monitor the execution status of the assigned tasks.
- Operational Resource Holon (ORH): is a physical entity within the manufacturing system, it can represent a robot, machine, worker, etc. The ORH is usually composed of two components. The first component is the physical component which represents the physical input/output (I/O) of a resource. The second component is the communication component which is responsible for translating the I/O events into information and conducting them to the other holons and vice-versa

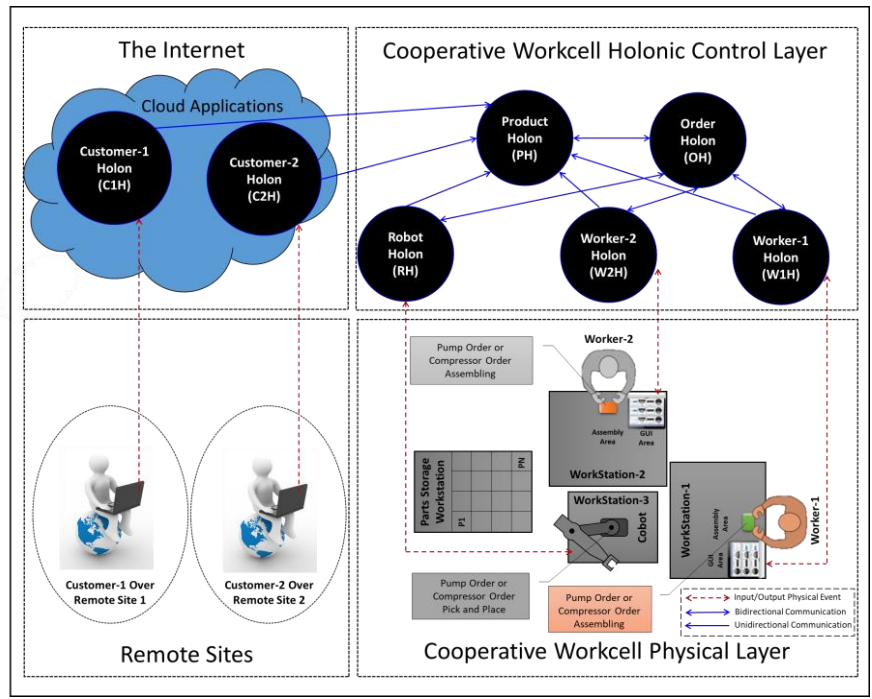

**Fig. 11.** Cooperative Assembly Workcell - a case study

Fig. 11 illustrates the idea of applying the HCA concept over a cooperative workcell which composes of two workers in cooperation with one cobot. The goal of the cooperative workcell is to assemble a family of a customized product. This case-study will be explained in details in the following section.

## 5 Case Study Implementation

### 5.1 Case Study Description

The goal of the case study is to implement the previously shown solution concept. Two Customer Holons (CHs) have been developed to customize the production orders as it can be seen in Fig. 12-a. Both the CHs have a similar UI. The UI of the CH is providing a tool for ordering a specific product with certain features (i.e., parts). The customer selects the basic features and defines the needed amount of the product then sends the order to the PH [2].

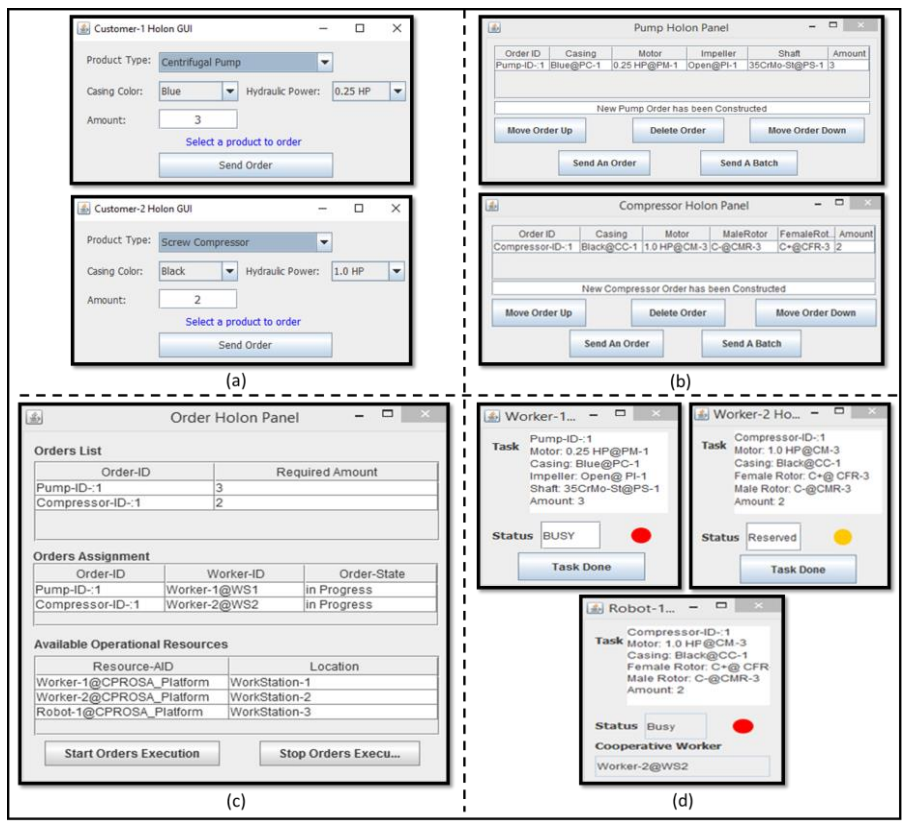

**Fig. 12.** (a) Customers Holon UI – (b) Products Holon UI – (c) Order Holon UI –

(d) Operation Resources Holon UI

Two products can be manufactured in this case-study, the first is a centrifugal pump and the second is a screw compressor. The UIs of the pump and the compressor holons can be seen in Fig. 12-b. The two products share some parts such as the casing and the electrical motor. The pump has two unique parts which are the impeller and the shaft, while the compressor has other two unique parts which are the male-rotor and the female-rotor. When a PH receives a product order from the CH, it constructs the building plans for this product order as it will be discussed later in details. The PH also has the ability to rearrange the orders or modify them before sending them to the OH. The OH is responsible for collecting the product orders from all the other PHs as it is shown in Fig. 12-c. Simultaneously the OH discovers the existence of the operation resources. Furthermore, it starts and stops the production process. Two WHs (W1H, W2H) and one Robot Holon (RH) can be found as operation resources in this implementation as it is shown in Fig. 12-d. The function of the workers within this case-study is to perform the assembly operation of the customized product orders, while the function of the cobot is to pick and place the customized parts of every production order to the proper worker workstation. As we do not have a robot hardware during this implementation, we assumed that the cobot always takes two seconds to pick and place one product. Therefore, the RH multiplies the number of products by two to obtain the overall time needed for the whole pick and place operations. Accordingly, the RH can have two statuses, either busy or free. Another status is required for the WH which is a reserve status. In the reserve status the WH is waiting the cobot to load at least one product to the worker, therefore the worker can start the assembly operation and subsequently the WH status turns to be busy. The WH stays in the busy status till the worker presses the task-done button, then the WH status would be free.

### 5.2 Case Study Ontology Model

As has been discussed earlier at the solution concept. JADE is using three different types of schemas to construct its ontology. Fig. 13 shows all the required schemas used to build the case-study ontology. The first set of schemas are the terms (i.e., concepts and primitives):

- Compressor-Customer-Order: a schema which encapsulates some attributes such as the required color, the needed hydraulic power, and the required amount. Also it contains an AID as every customer-order is an agent needs an ID.
- Pump-Customer-Order: a schema which encapsulates some attributes such as the required color, the needed hydraulic power, and the required amount. Also it contains an AID as every customer-order is an agent needs an ID.
- Casing: a shared part between the pump and the compressor. The casing schema contains two attributes which are the casing color, and its position at the storage workstation.
- Electrical-Motor: a shared part between the pump and the compressor. The motor schema contains two attributes which are the motor electrical power, and its position at the storage workstation.
- Shaft: a unique part of the pump. The shaft schema contains two attributes which are the shaft material, and its position at the storage workstation.
- Impeller: a unique part of the pump. The impeller schema contains two attributes which are the impeller type, and its position at the storage workstation.
- Female-Rotor: a unique part of the compressor. The female-rotor schema contains two attributes which are the rotor size, and its position at the storage workstation.

- Male-Rotor: a unique part of the compressor. The male-rotor schema contains two attributes which are the rotor size, and its position at the storage workstation.
- Compressor: a concept schema which encapsulates many other schemas under it, those schemas are the casing, electrical-motor, female-rotor, and male-rotor. Every compressor is an agent; therefore, it must contain an AID.
- Pump: a concept schema which encapsulates many other schemas under it, those schemas are the casing, electrical-motor, shaft, and impeller. Every pump is an agent; therefore, it must contain an AID attribute.
- Compressor-Order: a schema which extends the compressor schema by adding the required amount of units.
- Pump-Order: a schema which extends the pump schema by adding the required amount of units.
- Operations-List: a schema which includes a list of operations which can be used to manufacture either a pump or a compressor. The schema can be used to manufacture a product which needs three operations or less.
- Compressor-Manufacturing-Order: a schema which combines a Compressor-Order schema and an Operations-List schema. Also it has an AID attribute as it acts as an agent.
- Pump-Manufacturing-Order: a schema which combines a Pump-Order schema and an Operations-List schema. Also it has an AID attribute as it acts as an agent.
- Worker: a schema which contains two attributes, the first one is the worker AID as it acts as a life agent, and the second is the worker location within the workcell (i.e., workstation). The worker agent is providing an UI for the worker for providing the assigned task and inquiring the task done event (see Fig. 12-d). Two instances of the worker agent exist in this case-study scenario. The worker can have three statuses. A free status when there is no product orders or the production is not started. A reserve status when the worker is waiting the first product unit to be placed by the cobot. A busy status last while the cobot is handling the production orders till the worker triggers the task done event.
- Robot: a schema which contains one attribute, which is the robot AID as it acts as an agent. The robot schema does not have a workstation attribute because only there is only one cobot which is responsible for the pick and place. Therefore, the location of the cobot is not necessary required. However, in case of more than one cobot the cobot location attribute could be important. The robot agent provids an UI to show the assigned task and the status of the cobot (see Fig. 12-d). The cobot can have two statuses. A free status when there is no product orders or the production is not started. A busy status last while the cobot is handling the production orders. A timer of two second has been assigned to every pick and place operation

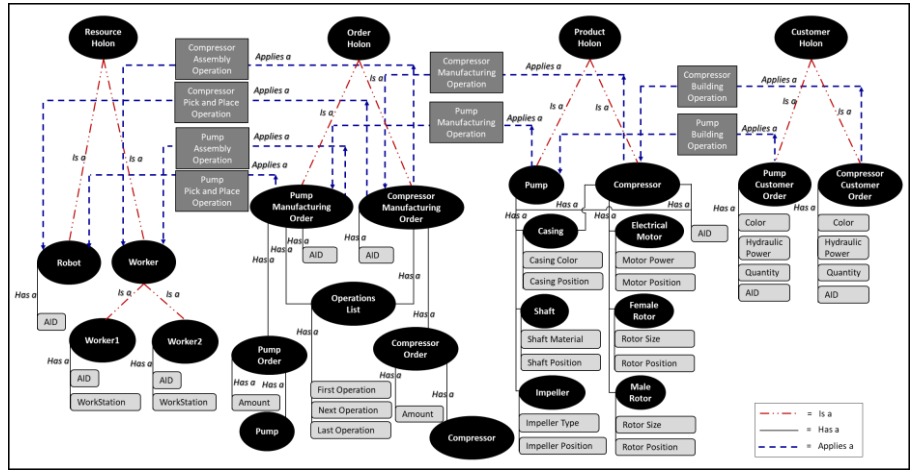

**Fig. 13.** Case-Study Ontology Model

The second set of schemas which can be seen in Fig. 13 are the predicate schemas which are addressed as the following:

- (concept-x) <Is-a> (concept-y): usually a relation between two concept schemas. This relation is similar to the object oriented abstraction. Thus, this predicate expression has been used to express the parent-child relationship between the concepts.
- (concept-x) <Has-a> (attribute-x): usually a relation between a concept and an attribute, an attribute can be a concept schema or a primitive. This relation is similar to object oriented inheritance. Thus, this predicate expression has been used to form sophisticated objects from simpler ones.
- (agent-x) <Applies-a> (action-x): usually a relation between a concept and an action schema. A concept uses this predicate expression to trigger one or more than one actions at the same time. The action schemas will be discussed below in details.

The third set of schemas which can be seen in Fig. 13 are the action schemas which are addressed as the following:

- Pump-Building-Operation: this action schema expects a Pump-Customer-Order concept schema as an input, and it is deployed either by customer-1 or customer-2 agents. An example of this operation can be seen at the ACL-Message content in Fig. 14-a.
- Compressor-Building-Operation: this action schema expects a Compressor-Customer-Order concept schema as an input, and it is deployed by either customer-1 or customer-2 agents. An example of this operation can be seen at the ACL-Message content in Fig. 14-b.
- Pump-Manufacturing-Operation: this action schema expects a Pump-Order and a Pump-Operations-List concept schemas as inputs, and it is deployed by the pump agent. A detailed example of this operation can be seen at the ACL-Message content in Fig. 14-c.
- Compressor-Manufacturing-Operation: this action schema expects a Compressor-Order and a Compressor-Operations-List concept schemas as inputs, and it is deployed by the compressor agent. A detailed example of this operation can be seen at the ACL-Message content in Fig. 14-d.

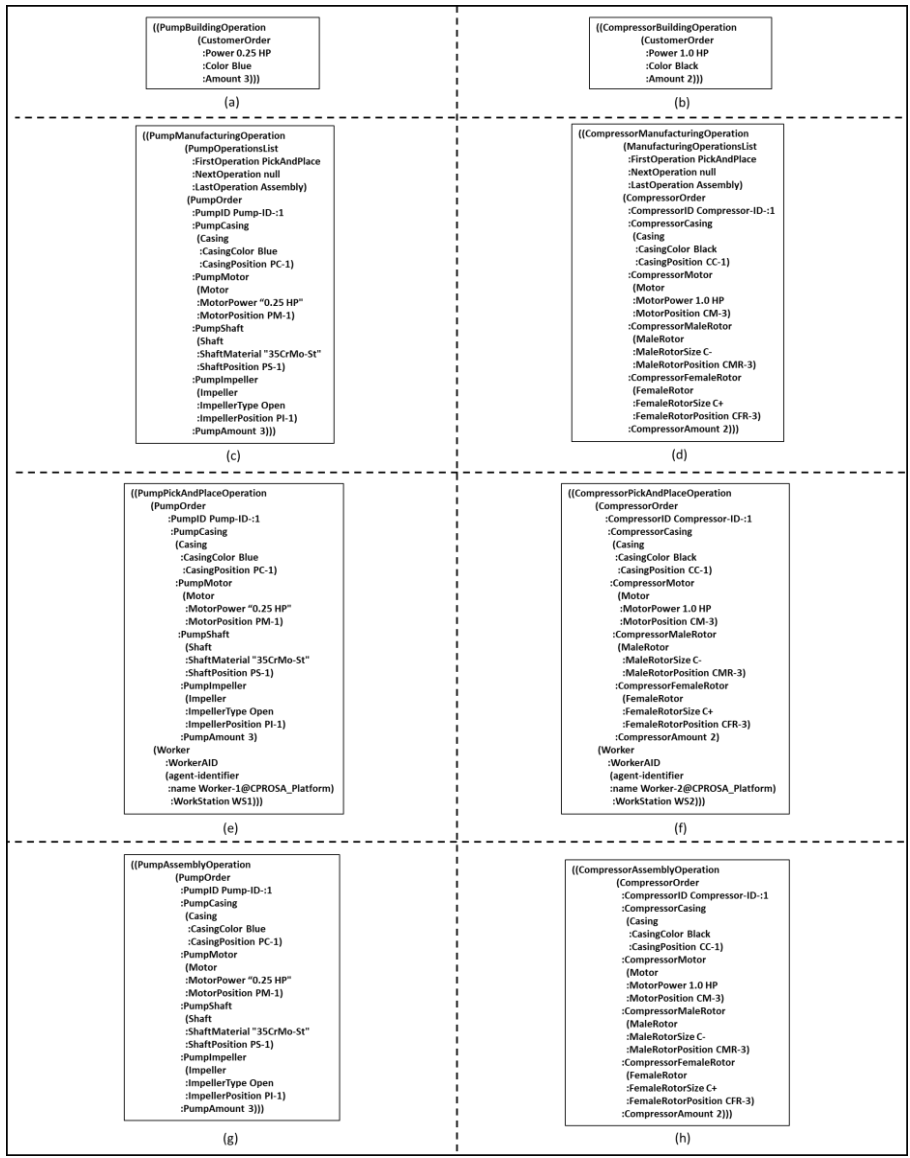

**Fig. 14.** Ontology-Based ACL-Messages used during the Case-Study

- Pump-Pick-And-Place-Operation: this action schema expects two inputs. The first input is the Pump-Order concept schema which contains the detailed specifications of the pump order. Therefore, the cobot uses this information especially the pump parts positions to perform the pick operation. The second input is the target worker concept schema. Therefore, the cobot uses the worker workstation location to place the pump parts at this location. This action schema is deployed by the orders agent to assign a task to the robot agent. A detailed example of this operation can be seen at the ACL-Message content in Fig. 14-e.
- Compressor-Pick-And-Place-Operation: this action schema expects two inputs. The first input is the Compressor-Order concept schema which contains the detailed specifications of the compressor order. Therefore, the cobot uses this information especially the compressor parts positions to perform the pick operation. The second input is the target worker concept schema. Therefore, the cobot can use the worker workstation location to place the compressor parts at this location. This action schema is deployed by the orders agent to interact with the robot agent. A detailed example of this operation can be seen at the ACL-Message content in Fig. 14-f.
- Pump-Assembly-Operation: this action schema expects one concept schema input which is the pump-order. This operation is beneficial for the worker to provide him with the required parts to build a customized pump. Moreover, it provides the amount of required units. This action schema is deployed by the orders agent to assign a task to any of the worker agents based on their status. A detailed example of this operation can be seen at the ACL-Message content in Fig. 14-g.
- Compressor-Assembly-Operation: this action schema expects one input which is a compressor-order concept schema. This operation is beneficial for the worker to provide the knowhow of building a customized compressor. Moreover, it provides the amount of required units. This action schema is deployed by the orders agent to assign a task to any of the worker agents based on their status. A detailed example of this operation can be seen in Fig. 14-h.

### 5.3 Case Study Holons Interaction

Fig. 15-a shows JADE interaction scenario among the CHs (i.e., customer-1 agent and customer-2 agent) and the PHs (i.e., pump agent and compressor agent). In this scenario, customer-1 agent sends an ACL-Message with an AGREE communicative act. The AGREE-message contains a Pump-Building-Operation and a Pump-Customer-Order. The AGREE-message is received by the pump agent. Therefore, the pump agent confirms the receiving by sending back a CONFIRM-message to customer-1 agent. Simultaneously the pump agent constructs a pump instance based on the incoming customer-order. The same mechanism is used between cusomter-2 agent and the compressor agent to construct a new instance of a compressor associated with a customer-2 order. Fig. 15-b shows JADE interaction scenario between the PHs (i.e. pump agent and compressor agent) and the OH. This interaction is following the same mechanism used in before in Fig. 15-a, except that it replaces the AGREE-messages with a PROPAGATE-messages.

Fig. 15-c shows JADE interaction scenario between the OH and the ORHs (i.e., worker-1 agent, worker-2 agent, and robot agent). During this interaction, the manufacturing operations are assigned to the operational resources based on their statuses. As it can be seen in lines 1, 2, 3, and 4 of Fig. 15-c, the orders agent sends two REQUEST-messages which are replied by two CONFIRM-messages. The first REQUEST-message assigns a Pump-Pick-And-Place-Operation to the robot agent. The second REQUEST-message assigns a Pump-Assembly-Operation to worker-1 agent. The reason that the pump-order has been processed first by the orders agent is that it is the first product order at the order list (refer to Fig. 12-c). In line 5 of Fig. 15-c, the robot agent sends an INFORM-REF-message to worker-1 agent to tell that it placed the first pump unit. Then, the robot agent sends two INFORM-IF-messages to the orders agent and worker-1 agent to tell that it finished handling all the required pump amounts (i.e., three pump units by referring to Fig. 12-a, b, and d). The two INFORM-IF-messages can be seen in lines 6, and 7 of Fig. 15-c. The same interaction mechanism can be seen in lines 9,10,11,12, and 13 to assign the compressor-order manufacturing operations to the worker-2 agent and the robot agent. Lines 14, and 15 of Fig. 15-c shows the INFORM-messages to express done-signals which are generated by worker-1 and worker-2 agents.

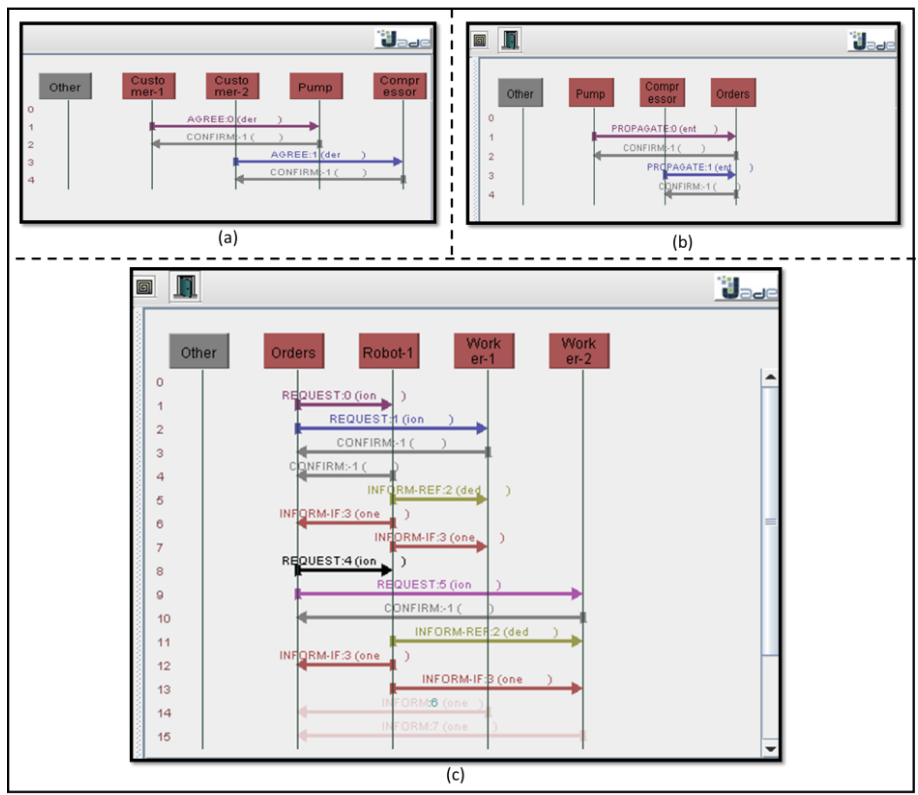

**Fig. 15.** Case-Study Interaction Scenario

# 6 Conclusion

This research has highlighted an important challenge in cooperative manufacturing. This challenge is the representation of the shared knowledge and the production entities in the cooperative workcell. Moreover, exchanging the represented manufacturing knowledge to obtain collective decisions. A holonic control concept has been purposed to solve this challenge. The implementation of this concept has been achieved via combining the autonomous agent communication with the ontology model. A case study of two workers in cooperation with one cobot has been chosen to verify the solution feasibility.

The conducted research presented the different approaches to implement an ontology model. Furthermore, implementing this model via the autonomous agent technology. The implementation showed that exchanging the manufacturing knowledge among the deployed agents supported the common understanding of the manufacturing environment and operations. By providing a natural language that can be comprehended by both the worker and the cobot.

Furthermore from the technical point of view, the implemented model is a practical solution to achieve the syntactic and structural interoperability. The same ontology can be re-applied over similar domains. For instance, the same model in the example can be applied over another cooperative workcell with the exact resource holon, with considering of changing the attribute values. Also, the same model can be extended to fit a cooperative workcell with more operational resources. Furthermore, this ontology model can be fused with another model to fit in a more sophisticated manufacturing scenario.